# Variational Bayesian modelling of mixed-effects

*J. Daunizeau*[1,2]

[1] Brain and Spine Institute, Paris, France

[2] ETH, Zurich, France

Address for correspondence:

Jean Daunizeau

Motivation, Brain and Behaviour Group

Brain and Spine Institute (ICM), INSERM UMR S975.

47, bvd de l'Hopital, 75013, Paris, France.

Tel: +33 1 57 27 43 26

Fax: +33 1 57 27 47 94

Mail: jean.daunizeau@gmail.com

Web: https://sites.google.com/site/jeandaunizeauswebsite/

Setting the priors is arguably one of the most delicate issues of Bayesian inference (Berger, 1985; Datta and Ghosh, 1995; Jeffreys and S, 1946; Robert, 2001). Although only flat priors are valid from a frequentist perspective, they can in fact be considered largely suboptimal, when compared to almost *any* form of informative prior (Gigerenzer and Brighton, 2009; Golchi, 2016). The reason lies in the so-called "bias-variance trade-off" of statistical estimation (Geman et al., 1992; Geurts, 2005): the systematic bias that may be induced by informative priors is overcompensated by the reliability of regularized parameter estimates. In brief, if one really cares about expected estimation and/or prediction error, then one should not aim for unbiasedness…

But acknowledging the benefit of priors does not solve the issue of setting them, in the commonplace situation where one does not have much solid ground to lay on. This is where so-called empirical Bayes methods may be useful (Carlin et al., 2000). In brief, these are procedures for statistical inference in which the prior distribution is estimated from the data. This approach stands in contrast to standard Bayesian methods, where the prior distribution is fixed before any data are observed. In this note, we will be concerned with a specific subcase of empirical bayes, which arises in the context of group studies, i.e. empirical studies that report multiple measurements acquired in multiple subjects. In brief, within-subject priors can be learned from estimates of the group distribution of effects of interest. This class of statistical analysis is called "mixed-effect" modelling (McCulloch and Neuhaus, 2005). When approached from a bayesian perspective, it typically relies upon a hierarchical generative model of the data, whereby both within- and between-subject effects contribute to the overall observed variance.

In what follows, we derive a simple variational bayesian (Beal, 2003; Blei et al., 2017) scheme for the treatment of mixed-effects. In particular, we will consider within-subject generative models that can be nonlinear, and hence rely on joint "mean field"/Laplace approximations to variational bayesian inference (Daunizeau, 2017; Friston et al., 2007).

Let $n$ be the number of subjects and $y_i$ be a $p \times 1$ vector of observations or samples for subject $i$, where $p$ is the number of sample per subject. We assume it can be described using the following generative model:

$$\begin{aligned} y_i &= g(\theta_i) + \varepsilon_i \\ \theta_i &= \nu + \eta_i \end{aligned} \quad (1)$$

where is the observation function, $\varepsilon_i \sim N(0, \sigma_i^{-1} Q_y)$ are i.i.d. gaussian within-subject residuals (with subject-dependent precision[1] $\sigma_i$), $\theta_i$ are $n_\theta \times 1$ vectors of subject-specific parameters, $\nu$ is the population mean ($n \times 1$ vector) and $\eta_i \sim N(0, \Lambda^{-1})$ are i.i.d. gaussian between-subject residuals (and $\Lambda = Diag\left(\begin{bmatrix} \lambda_1 & \lambda_2 & \cdots & \lambda_{n_\theta} \end{bmatrix}\right)$ is the parameter-specific between-subject precision matrix). Remark: in this model, we consider that parameters do not covary with each other at the group level. In addition, the matrix $Q_y$ is a known prior covariance structure (it will not be updated by the VB inference machinery). In particular, it can be used to remove a few data points ($Q_y \to \infty$). In contradistinction, the between-subject precisions $\lambda$ will be updated (see below). Note that setting $\lambda \to \infty$ effectively fix the corresponding parameter to the population mean, which is essentially equivalent to a fixed-effect analysis. Alternatively, inverting the model under the constraint that the group mean is zero ($v=0$), and with a unitary sample size ($n=1$), reduces to ARD (Automatic Relevance Determination) schemes, which provide sparse estimators for model parameters. We will comment further on these special cases below.

---

[1] precision here refers to inverse variance.

The corresponding generative model is summarized on Figure 1 below.

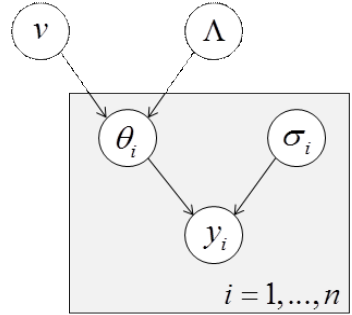

**Figure 1:** hierarchical model underlying mixed-effects analysis. The plate denotes a repetition over subjects ( $i=1,...,n$ ) in a group of sample size $n$

Note that classical random-effect analysis (RFX) can be performed using a summary statistics approach, which essentially bypasses this generative model and simply reports the within-subject parameter estimates $\hat{\theta}_i$ at the group level. This is, however, not optimal whenever the sample size is small or in the presence of heteroscedasticity. Alternatively, one may want to rely on such hierarchical model to define priors over model parameters that are empirically learned from group statistics. In what follows, we propose a VB approach to the full mixed-effect model, which inverts the above hierarchical model, while properly accounting for within- and between-subject variability.

In conjunction with statistical assumptions regarding within- and between-subjects residuals, Equation 1 induces the following joint distribution over variables:

$$p(y,\theta,\sigma,v,\lambda|m) = p(v|m) p(\lambda|m) \prod_{i=1}^{n} p(y_i|\theta_i,\sigma_i,m) p(\theta_i|v,\lambda,m) p(\sigma_i|m) \qquad (2)$$

where $p(v|m)$, $p(\lambda|m)$ and $p(\sigma_i|m)$ are the prior densities over the population mean, the population inverse variance and the within-subject residual precision. Without loss of generality, we will assume Gaussian priors on the population mean, i.e.: $p(v|m) = N(\mu_v^0, \Sigma_v^0)$, where $\mu_v^0$ and $\Sigma_v^0$ are known prior mean and variance matrix, respectively. Similarly, we assume Gamma priors for $\alpha$ and $\lambda$, as follows: $p(\lambda|m) = \prod_{j=1}^{n_\theta} Ga(a_\lambda^0, b_\lambda^0)$ and $p(\sigma|m) = \prod_{i=1}^{n} Ga(a_\sigma^0, b_\sigma^0)$, where prior scale and shape parameters do not depend upon subjects.

VB now proceeds from a mean-field separability assumption between $\theta = \{\theta_1, \theta_2, ..., \theta_n\}$, $\sigma = \{\sigma_1, \sigma_2, ..., \sigma_n\}$, $v$ and $\lambda = \{\lambda_1, \lambda_2, ..., \lambda_{n_\theta}\}$. In turn, the approximate posterior densities can be written as follows:

$$q(\theta) = \frac{1}{Z_\theta} \exp\langle \ln p(y,\theta,\sigma,v,\lambda|m)\rangle_{q(v)q(\alpha)q(\lambda)}$$

$$q(\sigma) = \frac{1}{Z_\sigma} \exp\langle \ln p(y,\theta,\sigma,v,\lambda|m)\rangle_{q(v)q(\lambda)q(\theta)}$$

$$q(v) = \frac{1}{Z_v} \exp\langle \ln p(y,\theta,\sigma,v,\lambda|m)\rangle_{q(\theta)q(\lambda)q(\sigma)}$$

$$q(\lambda) = \frac{1}{Z_\lambda} \exp\langle \ln p(y,\theta,\sigma,v,\lambda|m)\rangle_{q(v)q(\sigma)q(\theta)}$$

(3)

We will see that the mean-field approximation effectively decouples the different levels of the hierarchy, allowing for an efficient and simple message passing scheme using moments of the relevant approximate posterior distributions. In turn, stochastic dependencies between variables are replaced by deterministic dependencies between the moments of their respective posterior distributions.

We will now describe the VB approach to mixed-effects analysis, and highlight how this can be implemented using non-hierarchical model inversion.

### 1. Updating the within-subject effects

Let us first expose the derivation of the joint posterior density $q(\theta)$ over within-subject effects, which derives from decomposing the first line of Equation 3:

$$\begin{aligned}\ln q(\theta) &= -\ln Z_\theta + \langle \ln p(y,\theta,\sigma,v,\lambda|m)\rangle_{q(v)q(\lambda)q(\sigma)} \\ &= cst - \frac{1}{2}\sum_{i=1}^n \langle\sigma_i\rangle(y_i - g(\theta_i))^T Q_y^{-1}(y_i - g(\theta_i)) - \frac{1}{2}\sum_{i=1}^n (\theta_i - \langle v\rangle)^T \langle\Lambda\rangle(\theta_i - \langle v\rangle) \\ &= cst + \sum_{i=1}^n \ln q_i(\theta_i)\end{aligned}$$

(4)

where marginal posterior distributions $q_i(\theta_i)$ over within-subject effects are such that:

$$\ln q_i(\theta_i) = cst - \frac{1}{2}\langle\sigma_i\rangle(y_i - g(\theta_i))^T Q_y^{-1}(y_i - g(\theta_i)) - \frac{1}{2}(\theta_i - \langle v\rangle)^T \langle\Lambda\rangle(\theta_i - \langle v\rangle)$$

(5)

Thus, the above mean-field assumption implies the separability of $q(\theta)$ over subjects. This means that, given the posterior mean $\langle\Lambda\rangle$ and $\langle v\rangle$ of the two moments of the population distribution (as well as posterior means $\langle\sigma_i\rangle$ of the within-subject residual variances), the VB update of the marginal posterior density over within-subject effects can proceed for each subject independently of each other.

Let us now expose the derivation of the joint posterior density $q(\sigma)$ over within-subject residuals' precisions, which derives from decomposing the second line of Equation 3:

$$\ln q(\sigma) = -\ln Z_\sigma + \left\langle \ln p(y,\theta,\sigma,v,\lambda|m) \right\rangle_{q(v)q(\alpha)q(\theta)}$$

$$= cst + \frac{p}{2}\sum_{i=1}^{n}\ln \sigma_i - \frac{1}{2}\sum_{i=1}^{n}\sigma_i \left\langle (y_i - g(\theta_i))^T Q_y^{-1}(y_i - g(\theta_i)) \right\rangle + (a_\sigma^0 - 1)\sum_{i=1}^{n}\ln \sigma_i - \sum_{i=1}^{n}b_\sigma^0 \sigma_i \quad (6)$$

$$= cst + \sum_{i=1}^{n}\ln q_i(\sigma_i)$$

where marginal posterior distributions $q_i(\sigma_i)$ over within-subject residual precisions are such that:

$$\ln q_i(\sigma_i) = cst + \left(\frac{p}{2} + a_\sigma^0 - 1\right)\ln \sigma_i - \left(\frac{1}{2}\left\langle (y_i - g(\theta_i))^T Q_y^{-1}(y_i - g(\theta_i)) \right\rangle + b_\sigma^0\right)\sigma_i \quad (7)$$

Here again, $q(\sigma)$ factorizes over subjects.

Taken together, Equations 4-7 imply that the VB update of $q(\theta)$ and $q(\sigma)$ is equivalent to a subject by subject model inversion, whereby each within-subject model uses the same "effective" prior $\tilde{p}(\theta_i)$ over $\theta_i$, namely: $\tilde{p}(\theta_i) = N(\langle v \rangle, \langle \Lambda \rangle^{-1})$. This can be performed using any VB-Laplace machinery dealing with non-hierarchical model inversion techniques (see, e.g., (Daunizeau et al., 2014; Friston et al., 2007)), given the posterior estimates $\langle \Lambda \rangle$ and $\langle v \rangle$ of the two moments of the population distribution (see below). Suffice it to say that such within-subject VB inversion eventually yields sufficient statistics of the corresponding posterior distributions, i.e.: $q(\theta_i) = N(\mu_{\theta_i}, \Sigma_{\theta_i})$ and $q(\sigma_i) = Ga(a_{\theta_i}, b_{\theta_i})$, where: $\langle \theta_i \rangle \approx \mu_{\theta_i}$, $\langle \theta_{ij}\theta_{ij}^2 \rangle \approx \mu_{\theta_{ij}}^2 + \Sigma_{\theta_{ij}}$ and $\langle \sigma_i \rangle \approx a_{\sigma_i}/b_{\sigma_i}$.

In the next section, we expose the VB update of the moments of the population distribution, which now necessitates a more specific (though simple) procedure.

2. **Updating the between-subject effects**

Let us first expose the derivation of the posterior density $q(v)$ over the population mean, which derives from the third line of Equation 3, as follows:

$$\ln q(v) = -\ln Z_v + \left\langle \ln p(y,\theta,\sigma,v,\lambda|m) \right\rangle_{q(\theta)q(\lambda)q(\sigma)}$$

$$= cst - \frac{1}{2}\sum_{i=1}^{n}(\mu_{\theta_i} - v)^T \langle \Lambda \rangle (\mu_{\theta_i} - v) - \frac{1}{2}(\mu_v^0 - v)^T \Sigma_v^{0-1}(\mu_v^0 - v) \quad (8)$$

One can see from Equation 8 that the VB update of the population mean only requires the posterior means $\langle \theta_i \rangle$ of within-subject effects (which derives from the within-subject model inversions, see Equations 4-7 above), as well as the posterior mean $\langle \lambda \rangle$ of the population precision. One can easily show that the posterior density $q(v)$ is Gaussian, with moments given by:

$$q(v) = N(\mu_v, \Sigma_v) : \begin{cases} \mu_v = \Sigma_v \left( \Sigma_v^{0-1} \mu_v^0 + \langle \Lambda \rangle \sum_{i=1}^n \mu_{\theta_i} \right) \\ \Sigma_v = \left( \Sigma_v^{0-1} + n \langle \Lambda \rangle \right)^{-1} \end{cases} \quad (9)$$

At the uninformative prior limit (i.e. when $\Sigma_v^0 \to \infty$), Equation 9 simply reduces to: $\mu_v = 1/n \sum_{i=1}^n \mu_{\theta_i}$ and $\Sigma_v = \langle \Lambda \rangle^{-1}/n$.

Let us now expose the derivation of the posterior density $q(\lambda)$ over the population precision, which derives from the fourth line of Equation 3:

$$\begin{aligned}
\ln q(\lambda) &= -\ln Z_\alpha + \langle \ln p(y, \theta, \sigma, v, \lambda | m) \rangle_{q(\theta)q(v)q(\sigma)} \\
&= cst + \frac{n}{2} \ln |\Lambda| - \frac{1}{2} \sum_{i=1}^n \langle (\theta_i - v)^T \Lambda (\theta_i - v) \rangle + n_\theta \left( (a_\lambda^0 - 1) \ln \lambda_j - b_\lambda^0 \lambda_j \right) \\
&= cst + \frac{n}{2} \sum_{j=1}^{n_\theta} \ln \lambda_j - \frac{1}{2} \sum_{j=1}^{n_\theta} \lambda_j \sum_{i=1}^n \langle (\theta_{ij} - v_j)^2 \rangle + n_\theta \left( (a_\lambda^0 - 1) \ln \lambda_j - b_\lambda^0 \lambda_j \right) \\
&= \sum_{j=1}^{n_\theta} \ln q_j(\lambda_j)
\end{aligned} \quad (10)$$

where marginal posterior distributions $q_j(\lambda_j)$ over between-subject precisions are such that:

$$\ln q_j(\lambda_j) = cst + \left( \frac{n}{2} + a_\lambda^0 - 1 \right) \ln \lambda_j - \left( \frac{1}{2} \sum_{i=1}^n \langle (\theta_{ij} - v_j)^2 \rangle + b_\lambda^0 \right) \lambda_j \quad (11)$$

Here, $q(\lambda)$ factorizes over parameters, and each marginal posterior distribution $q_j(\lambda_j)$ has the form of a Gamma distribution, with shape and scale parameters given by:

$$q(\lambda_j) = Ga(a_{\lambda_j}, b_{\lambda_j}) : \begin{cases} a_{\lambda_j} = \frac{n}{2} + a_\lambda^0 \\ b_{\lambda_j} = \frac{1}{2} \sum_{i=1}^n \langle (\theta_{ij} - v_j)^2 \rangle + b_\lambda^0 \end{cases} \quad (12)$$

Equation 12 is the VB update equation for the approximate posterior density over $\lambda$. In particular, the expected population precision is given by: $\langle \lambda_j \rangle = a_{\lambda_j} / b_{\lambda_j}$. The expected precision matrix $\langle \Lambda \rangle$ is simply formed by diagonalizing the corresponding precision vector, i.e.: $\langle \Lambda \rangle = Diag\left( \begin{bmatrix} a_{\lambda_1}/b_{\lambda_1} & a_{\lambda_2}/b_{\lambda_2} & \cdots & a_{\lambda_{n_\theta}}/b_{\lambda_{n_\theta}} \end{bmatrix} \right) = A_\lambda / B_\lambda$, where $A_\lambda / B_\lambda$ is an abuse of notation. At first look, it may seem that 9 and 11 are, again, simple VB updates of an equivalent non-hierarchical GLM. This is not quite true however, because the VB update in Equation 11 requires both posterior means and variances of $v_j$ and $\theta_{ij}$:

$$\left\langle \left(\theta_{ij} - v_j\right)^2 \right\rangle = \left(\mu_{\theta_{ij}} - \mu_{v_j}\right)^2 + \Sigma_{v_j} + \Sigma_{\theta_{ij}} \qquad (13)$$

In particular, the last term in the right-hand term of Equation 13 accounts for estimation uncertainty on within-subject effects. Nevertheless, VB updates of the approximate posterior densities of the population mean and precision can be performed directly from the summary statistics of within-subject model inversions, according to Equations 8-13.

### 3. VB pseudo-code

The above derivations suggest the following pseudo-code for mixed-effects modelling in VBA:

---

*initialize* posterior $q(v)$ over population mean: $\mu_v \leftarrow \mu_v^0$ and $\Sigma_v \leftarrow \Sigma_v^0$.

*initialize* posterior $q(\lambda)$ over population precision: $a_{\lambda_j} \leftarrow a_\lambda^0$ and $b_{\lambda_j} \leftarrow b_\lambda^0$.

*until convergence*

    *for i=1:n (loop over subjects)*

        define within-subject priors as follows: $\tilde{p}(\theta_i) = N(\mu_v, A_\lambda / B_\lambda)$

        perform within-subject model inversion, i.e. update $q_i(\theta_i)$ and $q_i(\sigma_i)$

        store posterior summary statistics, i.e.: $\mu_{\theta_i}$, $\Sigma_{\theta_i}$, $a_{\sigma_i}$ and $b_{\sigma_i}$

    *end*

    update $q(v)$ and $q(\alpha)$ according to Equations 9 and 11-12.

    store posterior summary statistics, i.e.: $\mu_v$, $\Sigma_v$, $a_\lambda$ and $b_\lambda$

*end*

---

Convergence can be monitored using relative changes in the summary statistics of the approximate posterior densities. Alternatively, one can compute the free energy of the MFX model, as follows:

VB modelling of mixed-effects

$$F = \left\langle \ln p(y,\theta,\sigma,v,\lambda|m) \right\rangle_{q(\theta)q(v)q(\lambda)q(\sigma)} + S(q(\theta)) + S(q(v)) + S(q(\lambda)) + S(q(\sigma))$$

$$= -\frac{np}{2}\ln 2\pi + \frac{p}{2}\sum_{i=1}^{n}\langle \ln \sigma_i \rangle - \frac{1}{2}\sum_{i=1}^{n}\langle \sigma_i \rangle \left\langle (y_i - g(\theta_i))^T (y_i - g(\theta_i)) \right\rangle$$

$$+ (a_\sigma^0 - 1)\sum_{i=1}^{n}\langle \ln \sigma_i \rangle - b_\sigma^0 \sum_{i=1}^{n}\langle \sigma_i \rangle + n\left(a_\sigma^0 \ln b_\sigma^0 + \ln \Gamma(b_\sigma^0)\right)$$

$$-\frac{n n_\theta}{2}\ln 2\pi + \frac{n}{2}\sum_{j=1}^{n_\theta}\langle \ln \lambda_j \rangle - \frac{1}{2}\sum_{j=1}^{n_\theta}\langle \lambda_j \rangle \sum_{i=1}^{n}\left\langle (\theta_{ij} - v_j)^2 \right\rangle$$

$$+ (a_\lambda^0 - 1)\sum_{j=1}^{n_\theta}\langle \ln \lambda_j \rangle - b_\lambda^0 \sum_{j=1}^{n_\theta}\langle \lambda_j \rangle + n_\theta \left(a_\lambda^0 \ln b_\lambda^0 + \ln \Gamma(b_\lambda^0)\right)$$

$$-\frac{n_\theta}{2}\ln 2\pi - \frac{1}{2}\ln|\Sigma_v^0| - \frac{1}{2}(\mu_v^0 - \mu_v)^T \Sigma_v^{0-1}(\mu_v^0 - \mu_v) - \frac{1}{2}tr\Sigma_v^{0-1}\Sigma_v^0$$

$$+ S(q(\theta)) + S(q(v)) + S(q(\lambda)) + S(q(\sigma))$$

$$= \sum_{i=1}^{n} F_i$$

$$-\frac{n}{2}\sum_{j=1}^{n_\theta}\ln\langle \lambda_j \rangle + \left(a_\lambda^0 + \frac{n}{2} - 1\right)\sum_{j=1}^{n_\theta}\langle \ln \lambda_j \rangle - \sum_{j=1}^{n_\theta}\left(\frac{n}{2}\Sigma_{v_j} + b_\lambda^0\right)\langle \lambda_j \rangle + n_\theta\left(a_\lambda^0 \ln b_\lambda^0 + \ln \Gamma(b_\lambda^0)\right)$$

$$-\frac{n_\theta}{2}\ln 2\pi - \frac{1}{2}\ln|\Sigma_v^0| - \frac{1}{2}(\mu_v^0 - \mu_v)^T \Sigma_v^{0-1}(\mu_v^0 - \mu_v) - \frac{1}{2}tr\Sigma_v^{0-1}\Sigma_v^0 \qquad (14)$$

$$+ S(q(\lambda)) + S(q(v))$$

where $F_i$ is the subject-specific free energy derived from within-subject VB model inversions. One can see that the MFX free energy is the sum over within-subject free energies, plus a correction term that accounts for uncertainty regarding the population mean and variances.

4. Modelling fixed-effects

As we highlighted above, one can use this procedure to control the amount of expected between-subject variance using priors, eventually reducing the analysis to fixed-effects.

For example, let us assume that a given parameter (a pre-specified entry $\theta_{\cdot j}$ of the vector $\theta_{\cdot}$) is the same across subjects, i.e.: $\theta_{ij} = cst \ \forall i = 1,...,n$. In principle, this can be modelled by taking the limit $\lambda_j \to \infty$. However, this breaks the mean-field assumption, because the posterior over the corresponding parameter should now directly map to the population mean.

In fact, in this case, the VB update of $q(v)$ should be modified as follows:

$$q(v) = N(\mu_v, \Sigma_v) : \begin{cases} \mu_v = \Sigma_v \sum_{i=1}^{n} \Sigma_{\theta_i}^{-1} \mu_{\theta_i} \\ \Sigma_v = \left( \sum_{i=1}^{n} \Sigma_{\theta_i}^{-1} \right)^{-1} \end{cases} \quad (15)$$

where the corresponding prior over $\theta_i$ had been set to $\tilde{p}(\theta_i) = N(\mu_v^0, n\Sigma_v^0)$. This derives from the recursive application of Bayes theorem over subjects, having accounted for the multiple use of the prior density over model parameters. In fact, one also needs to add in a simple correction term to the within-subject free energies, i.e.:

$$F_i \leftarrow F_i + \frac{(n-1)n_\theta^{FFX}}{2} \ln 2\pi \quad (16)$$

where $n_\theta^{FFX}$ is the number of parameters that are assumed to be fixed-effects.

Note that such fixed-effect prior is different from assuming that the corresponding population mean is known (by zeroing the corresponding diagonal entry in the prior covariance matrix $\Sigma_v^0$). Only when the two assumptions are used in conjunction, would the scheme fix the parameter to its priors value $\mu_v^0$ during the VBA within-subject model inversion.

## 5. Concluding remarks

In this note, we have described a simple VB approach to mixed-effects modelling, which can jointly account for both random and fixed effects. We note that an open-source implementation of it is available as part of the VBA toolbox (Daunizeau et al., 2014): https://mbb-team.github.io/VBA-toolbox/.

The above VB approach to mixed effects is essentially an empirical Bayes approach to model inversion. It can be used to somehow bypass the specification of prior distributions for unknown model parameters, which is iteratively refined and converges toward the group distribution. Strictly speaking however, this procedure is valid only under the assumption that the same model (namely: $m$) underlies the generation of observed data for all subjects. In other words, it neglects the possibility that models may vary across subjects (Rigoux et al., 2014; Stephan et al., 2009). This raises the question of how to extend this approach to account for random effects both at the model and at the parameter level. We will briefly discuss a potential solution to this problem.

Let $m_i$ be a $K \times 1$ multinomial vector that encodes the identity of the model that captures subject $i$'s observed data. Following previous work on group-level model comparison, one can assume that the parent population can be described in terms of its frequency profile $r$ of models, where $r$ is a $K \times 1$ vector such that: $1 = \sum_{k=1}^{K} r_k$ and $0 \leq r_k \leq 1$. The resulting joint probability distribution over unknown variables now writes:

$$p(y,\theta,\sigma,\nu,\lambda,m,r) = p(m|r) p(r) \prod_{k=1}^{K} p(\nu_k|m_{ik}=1) p(\lambda_k|m_{ik}=1)$$

$$\times \prod_{i=1}^{n} p(y_i|\theta_{ik},\sigma_{ik},m_{ik}=1) p(\theta_{ik}|\nu_k,\lambda_k,m_{ik}=1) p(\sigma_{ik}|m_{ik}=1)$$

(17)

where all within- and between-subject variables are now indexed by their corresponding model index $k \in [1,K]$. This would induce modifications in the above VB modelling of mixed-effects, such that updates of within-subject and between-subject posterior densities on model parameters would deviate from priors in proportion to the posterior probabilities of model-subject assignments $p(m|y)$. The ensuing VB scheme would be different from first using an MFX analysis for each model separately, and then performing a RFX-BMS (Rigoux et al., 2014; Stephan et al., 2009) for selecting the models.